\documentclass{ceurart}

\usepackage{booktabs}
\usepackage{array}
\usepackage{url}
\usepackage{xcolor}
\usepackage{listings}
\usepackage{graphicx}
\usepackage{float}
\begin{document}

\copyrightyear{2026}
\copyrightclause{Copyright for this paper by its authors. Use permitted under Creative Commons License Attribution 4.0 International (CC BY 4.0).}
\conference{IberLEF 2026, September 2026, Le\'on, Spain}

\title{HULAT2 at MER-TRANS 2026: Governed Multi-Agent Simplification for Spanish Easy-to-Read Generation}

\address[1]{Computer Science and Engineering Department, Universidad Carlos III de Madrid, Av. Universidad, 30, Legan\'es, 28911, Spain}

\author[1]{Lourdes Moreno}[
orcid=0000-0002-9021-2546,
email=lmoreno@inf.uc3m.es,
url=https://hulat.inf.uc3m.es/
]

\author[1]{Paloma Mart\'inez}[
orcid=0000-0003-3013-3771,
email=paloma.martinez@uc3m.es,
url=https://hulat.inf.uc3m.es/
]
\author[1]{Marco Antonio Sanchez-Escudero}[
orcid=0009-0001-8163-5440,
email=marcoasa@inf.uc3m.es
]
\author[1]{Miguel Domínguez-Gómez}[
orcid=0009-0001-8163-5440,
email=100451258@alumnos.uc3m.es
]

\begin{abstract}
This paper describes the participation of HULAT2-UC3M in the Spanish track of MER-TRANS 2026, a shared task on multilingual Easy-to-Read translation. Three fully automatic Spanish runs were submitted. RUN1 and RUN2 used a LangGraph-based multi-agent workflow combining Gemini 2.5 Flash and RigoChat-7B-v2, parallel generation strategies, internal quality signals, Event--Condition--Action routing, controlled editing and traceable decisions. RUN1 used the base workflow, while RUN2 activated an additional lexical-support layer based on a glossary and lexical resources. RUN3 was a RigoChat-based generate--evaluate--regenerate baseline with prompt engineering and LoRA-based adaptation.

The official leaderboard reports BLEU-Orig, BLEU-Gold, SARI and BERTScore. During development, additional internal signals were also inspected, including semantic fidelity, readability, lexical simplicity, syntactic clarity and factual consistency. According to official SARI, RUN1 was the best HULAT2 run, with 44.0543 points, followed by RUN2 with 43.1049 and RUN3 with 38.5136. These results indicate that, in this task setting, signal-guided multi-agent routing outperformed the linear regeneration baseline. They also show that adding lexical support did not automatically improve reference-based scores. Further segment-level and document-level analysis are required to assess readability, factual consistency and user-oriented adequacy.
\end{abstract}


\maketitle

\section{Introduction}\label{sec:introduction}

Access to understandable information is essential for participation, autonomy and access to public services. This is particularly relevant for people with cognitive accessibility needs, reading comprehension difficulties, low literacy, older adults, non-native speakers and users facing unfamiliar or specialised content. Cognitive accessibility is concerned with reducing barriers to understanding, communication and interaction, and has been formalised in standards such as ISO 21801-1:2020 \cite{iso21801}. In text-based communication, these barriers are addressed through approaches such as Plain Language (PL), Easy-to-Read (E2R) and automatic text simplification.

PL and E2R are related but not equivalent. PL focuses on making information clear, findable, understandable and usable for its intended audience, as defined by ISO 24495-1:2023 \cite{iso24495}. E2R usually applies stricter constraints on vocabulary, syntax, layout and validation, and is particularly oriented towards people with reading comprehension difficulties. In Spain, E2R practice is supported by UNE 153101:2018 EX, which provides recommendations for the production and validation of Easy-to-Read documents \cite{une153101}. International guidance, such as the IFLA guidelines and the Inclusion Europe recommendations, also emphasises the need for clear language, simple structure and reader-oriented validation \cite{ifla2010easyread,inclusioneurope2009easyread}. For this reason, accessibility-oriented simplification should not be reduced to lexical substitution or sentence shortening; it also involves content selection, explicit relations between ideas, document structure, control of terminology, and the preservation of relevant information.

Automatic text simplification has evolved from rule-based and statistical approaches to neural models and, more recently, Large Language Models (LLMs) \cite{saggion2017automatic,nisioi2017neural}. LLMs can perform lexical substitution, sentence splitting, paraphrasing, deletion of secondary information and explanatory rewriting. However, LLM-based simplification also introduces risks: outputs may be fluent and apparently simple while omitting relevant information, adding unsupported explanations, changing numerical information or altering the meaning of the source. This is especially relevant in accessibility-oriented settings, where the output must be not only simpler, but also reliable, usable and appropriate for the intended readers. In Spanish, previous work has explored decoder-based LLMs for Easy-to-Read generation using parallel E2R resources, fine-tuning and expert qualitative evaluation \cite{martinez2024exploring}. These results show the potential of LLM-based simplification, while reinforcing the need for terminology support and human validation.

Evaluation remains a central challenge. Metrics such as BLEU, SARI and BERTScore are useful for benchmarking, but they capture only partial aspects of quality \cite{papineni2002bleu,xu2016sari,zhang2019bertscore}. BLEU rewards surface overlap with references, SARI evaluates add, keep and delete operations, and BERTScore estimates semantic similarity through contextual embeddings. Previous work has shown that automatic metrics may penalise valid reformulations and do not fully capture whether an adapted text is clearer, better organised or more useful for its intended audience \cite{alvamanchego2021unsuitability,maddela2023lens}. Other work has also pointed out that sentence-level evaluation can be too narrow for document-level simplification and that simplicity and meaning preservation should be assessed as different dimensions \cite{maddela2025document,cripwell2024readi}. Human-centred evaluation studies further show that expert judgments, automatic scores and actual reader comprehension may diverge \cite{agrawal2023human,carrer2024multifaceted}.

Recent workshops and shared tasks have contributed to more systematic evaluation settings for simplification, readability and accessibility. The TSAR series has consolidated research on text simplification, accessibility and readability \cite{saggion2022tsar,tsar2023,tsar2024,tsar2025}. TSAR 2025 included a shared task on readability-controlled text simplification, where systems had to adapt texts to specified CEFR-based proficiency levels while preserving meaning and fluency \cite{alvamanchego2025tsar}. READI 2024 addressed tools and resources for people with reading difficulties and included work on document-level simplification and evaluation beyond isolated sentences \cite{readi2024,cripwell2024readi}. In the Spanish context, CLEARS 2025 addressed PL and E2R adaptation within IberLEF, using semantic similarity and the Fern\'andez-Huerta readability index as official metrics \cite{botellagil2025clears}. These initiatives show a shift from generic simplification towards controlled, multilingual, language-specific and accessibility-oriented evaluation settings. They also highlight the difficulty of balancing semantic preservation, readability, user-oriented adequacy and reference alignment.

MER-TRANS 2026 extends this line of work to multilingual Easy-to-Read translation. The task asks participating systems to generate E2R-oriented versions of complex texts, with official results reported using BLEU-Orig, BLEU-Gold, SARI and BERTScore \cite{saggion2026mertrans}. The contribution of this paper is to describe the HULAT2-UC3M Spanish submissions to MER-TRANS 2026, using a signal-guided multi-agent architecture for controlled E2R-oriented generation, and to analyse the official results.

\section{Task and Data}\label{sec:task_data}

MER-TRANS 2026 was organised as part of IberLEF 2026
\cite{saggion2026mertrans,iberlef2026overview}.
The shared task uses data resources covering multilingual
Easy-to-Read adaptation and sentence difficulty assessment. The iDEM
corpus provides human-annotated original and Easy-to-Read texts designed
to support access to democratic participatory processes \cite{bott2026idem}.
The shared-task documentation also includes the Arabic sentence
difficulty classification dataset introduced by Khallaf and Sharoff
\cite{khallaf-sharoff-2021-automatic}.

MER-TRANS 2026 evaluates automatic systems for generating Easy-to-Read
(E2R) adaptations from original complex texts. HULAT2-UC3M submitted
only Spanish runs, and no human intervention was used during official
inference.

The MER-TRANS evaluation materials describe several metric families for the shared task, including surface-similarity metrics, simplification-oriented metrics, semantic-similarity metrics, readability metrics and complexity classifiers \cite{mertrans2026website}. However, the official evaluator repository used for the shared task computes BLEU, SARI and BERTScore, and the published leaderboard reports BLEU against the original text (BLEU-Orig), BLEU against the gold E2R reference (BLEU-Gold), SARI and BERTScore \cite{mertrans2026evaluator,mertrans2026website}. Therefore, official results are reported in this paper using these leaderboard metrics.

SARI is especially relevant for simplification because it evaluates add, keep and delete operations with respect to the source and the reference \cite{xu2016sari}. BLEU provides a surface-overlap signal, while BERTScore provides a complementary semantic-similarity signal based on contextual embeddings \cite{zhang2019bertscore}. These metrics offer a shared benchmark for the task, but they do not fully capture accessibility-oriented quality dimensions such as factual consistency, terminology support, readability, document coherence or user-oriented adequacy.

For development-time evaluation, the HULAT2 systems were not configured exclusively to maximise the official scores. SARI, BLEU and BERTScore were used as external benchmarking indicators, while candidate generation, routing and retry decisions were guided by a broader set of internal quality signals, including simplification, semantic-similarity, factual-consistency and readability indicators \cite{hulat2026bilingualmetrics}. In the MER-TRANS experiments, these development-time signals were used to support prompt calibration, Event-Condition-Action (ECA) routing, retry conditions and qualitative error analysis.

\section{System Description}\label{sec:system}

This section describes the resources, empirical evidence base and system configurations used for the HULAT2-UC3M Spanish submissions to MER-TRANS 2026. The section first presents the resources used during development, then summarises the submitted runs, and finally describes the multi-agent workflow and the baseline system.

\subsection{Resources and Evidence Base}\label{subsec:resources}

For system calibration, prompt refinement and glossary construction, publicly available Spanish E2R materials related to institutional, civic and public information contexts were used. These resources supported the analysis of sentence segmentation patterns, explanatory formulations, lexical choices and recurrent E2R conventions. They included \textit{La Constituci\'on Espa\~nola en lectura f\'acil}, citizen participation guides, electoral participation materials, and opposition-exam preparation materials published by Plena Inclusi\'on and public administrations. A detailed list of the public resources used for calibration and lexical support is provided in Appendix~\ref{app:public_resources}.

Part of these materials was used to build an internal HULAT-UC3M glossary of learning words and short explanatory forms. The glossary was used as lexical support in RUN2. RUN2 also incorporated Spanish medical lexical simplification resources \cite{campillos2022clara}. Although the main MER-TRANS domain is not medical, these resources were useful as additional lexical-simplification material for difficult terms and as explanatory paraphrases. The resources were not part of the official MER-TRANS test set.

In addition to these textual resources, the system design was informed by an empirical evidence base previously compiled for accessible text generation. This evidence base includes results from user studies, annotated resources and accessibility-oriented evaluations, and was operationalised as quality signals, checklists, ECA rules and review criteria for accessible simplification \cite{moreno2026hitlhotl}. In the MER-TRANS runs, this evidence was not used as human intervention during official inference, but as development-time support for prompt adjustment, routing rules, retry conditions and qualitative analysis.

\subsection{Submitted Runs}\label{subsec:runs}

Three Spanish runs were submitted. RUN1 used the base LangGraph-based multi-agent workflow, with parallel generation, candidate evaluation, ECA-style routing and controlled editing. RUN2 used the same workflow, but activated the lexical agent before generation in order to test whether glossary-based and lexical-resource support improved the base configuration. RUN3 was a separate RigoChat-based generate--evaluate--regenerate baseline, designed to compare the multi-agent routing strategy with a more linear simplification pipeline.

Table~\ref{tab:components} summarises the submitted configurations at pipeline level. The table distinguishes the multi-agent workflow used in RUN1 and RUN2 from the linear baseline workflow used in RUN3, and specifies which stages were shared, optional or exclusive to each run.

\begin{table}[!htbp]
\centering
\caption{Pipeline-level description of the HULAT2-UC3M submitted configurations.}
\label{tab:components}
\small
\begin{tabular}{@{}p{0.15\linewidth}p{0.16\linewidth}p{0.20\linewidth}p{0.34\linewidth}p{0.07\linewidth}@{}}
\toprule
\textbf{Pipeline stage} & \textbf{Stage or module} & \textbf{Main resource or model} & \textbf{Function} & \textbf{Runs} \\
\midrule
\multicolumn{5}{@{}l}{\textit{Multi-agent workflow used in RUN1 and RUN2}} \\
\midrule
Optional lexical support & Lexical agent & Glossary and lexical resources & Detects difficult terms and proposes lexical support before generation. This stage is skipped when the lexical agent is disabled. & RUN2 \\
Pre-analysis & Pre-analysis module & Rule-based feature extraction & Detects structural, lexical and factual-risk features, including sentence length, punctuation complexity, numbers, dates, negations, conditions and glossary terms. & RUN1, RUN2 \\
Parallel generation & Generator A & Gemini 2.5 Flash & Produces a conservative Plain Language-oriented candidate, prioritising meaning preservation and minimal reformulation. & RUN1, RUN2 \\
Parallel generation & Generator B & RigoChat-7B-v2 & Produces a structurally simplified candidate, prioritising sentence splitting, directness and reduced lexical difficulty. & RUN1, RUN2 \\
Parallel generation & Generator C & Gemini 2.5 Flash & Produces an Easy-to-Read-oriented candidate guided by short sentences, common vocabulary, explicit structure and an operational CEFR A2-like target. & RUN1, RUN2 \\
Candidate assessment & Candidate evaluator & Internal quality signals & Assesses candidates using semantic, factual, syntactic, lexical, readability and robustness signals. & RUN1, RUN2 \\
Routing and control & ECA router & Event--Condition--Action rules & Selects, edits, merges or retries candidates according to the detected quality profile and risk signals. & RUN1, RUN2 \\
Controlled editing & Merger-editor & Restricted editing prompt & Applies minimal controlled edits or fuses candidates only when complementary improvements are detected without semantic-risk signals. & RUN1, RUN2 \\
Final validation & Final evaluator & Internal validation signals & Rechecks factual consistency, semantic preservation and robustness before producing the final output. & RUN1, RUN2 \\
\midrule
\multicolumn{5}{@{}l}{\textit{Linear baseline workflow used in RUN3}} \\
\midrule
Generation & Baseline generator & RigoChat-7B-v2 with prompt engineering and LoRA adaptation & Generates an Easy-to-Read-oriented output using a controlled baseline configuration. & RUN3 \\
Quality control & Baseline evaluator & Internal readability and semantic-preservation checks & Evaluates the generated output using internal quality criteria, including readability, semantic preservation and critical-error checks. & RUN3 \\
Regeneration & Regeneration loop & Regeneration prompt & Attempts regeneration when internal quality criteria are not met. & RUN3 \\
\bottomrule
\end{tabular}
\end{table}

\subsection{Multi-Agent Workflow}\label{subsec:multiagent}

RUN1 and RUN2 were generated with the same LangGraph-based multi-agent architecture. LangGraph was used to implement a stateful workflow with specialised nodes, conditional routing, retry loops and traceable intermediate decisions \cite{langgraph2026}. The system combines Gemini 2.5 Flash \cite{gemini25flash} and RigoChat-7B-v2 \cite{gomez2025rigochat}. RigoChat-7B-v2 is a Spanish-oriented model based on Qwen2.5-7B-Instruct and further adapted for Spanish queries.

\begin{figure}[t]
    \centering
    \includegraphics[width=\linewidth]{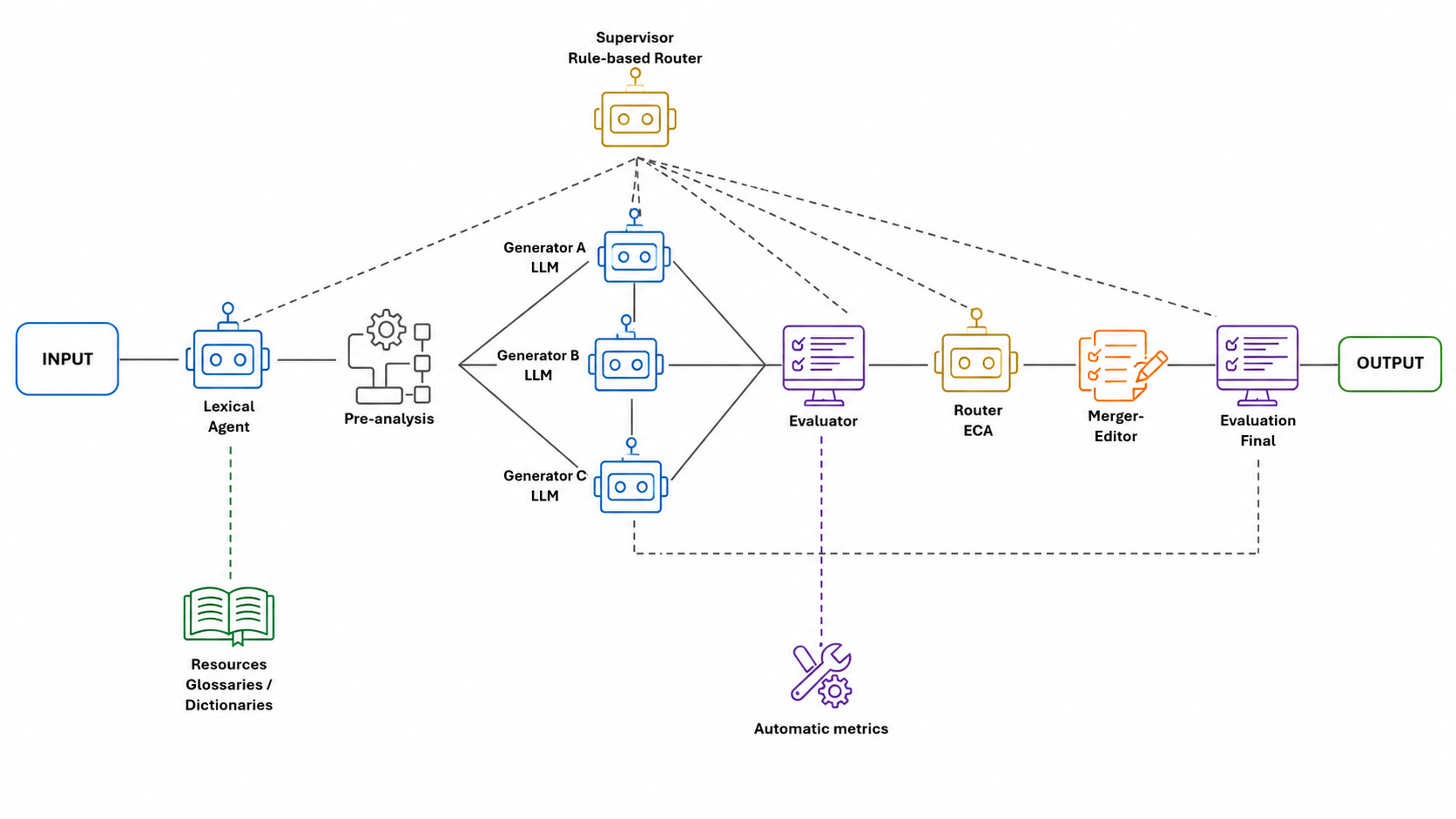}
    \caption{LangGraph-based multi-agent workflow used for RUN1 and RUN2 in the MER-TRANS 2026 Spanish submissions.}
    \label{fig:architecture}
\end{figure}

\paragraph{Workflow overview.}
Figure~\ref{fig:architecture} summarises the workflow used for the MER-TRANS 2026 Spanish submissions. A demonstration video illustrating the workflow execution and traceable decisions is available online.\footnote{\url{https://youtu.be/FV4bymTRcQo?si=dgYNKsc2j3lxXUP0}} The lexical agent is optional: it was disabled in RUN1 and enabled in RUN2. The figure also shows the role of the rule-based supervisor, the use of automatic evaluation signals and the controlled path from candidate generation to final output.

The workflow starts from the input text. When the lexical agent is enabled, lexical resources are consulted before pre-analysis; when it is disabled, the input is passed directly to the pre-analysis stage. The pre-analysis module characterises the input before generation by detecting structural, lexical and factual-risk features, including sentence length, punctuation complexity, numbers, dates, negations, conditions and possible glossary terms. These features are used to determine whether the text requires conservative rewriting, syntactic simplification, lexical support or stronger semantic control.

\paragraph{Evidence-informed signals and control rules.}

Some of these signals, rules and threshold ranges were informed by the empirical evidence base. In particular, the evidence supported the definition of quality priorities, checklists, ECA-style rules and acceptance, warning, retry and rejection criteria for accessible simplification \cite{moreno2026hitlhotl}. Other signals were defined as operational heuristics to control generation robustness, lexical difficulty, sentence structure and unsupported content during the MER-TRANS configuration.

In this configuration, high-risk semantic and factual signals were treated as stricter constraints, while syntactic, lexical and E2R-format signals were treated as calibrable criteria during development. The signal taxonomy, ECA rule specification and threshold configuration are documented in the accompanying material archived in Zenodo \cite{moreno2026aigovaccess}. In this paper, the operational signal categories and decision logic are reported at the level needed to interpret the submitted runs, while the extended technical specification is provided as supplementary material.

\paragraph{Parallel generation strategies.}

Three candidate generators are launched in parallel. Generator A uses Gemini 2.5 Flash with a conservative Plain Language-oriented strategy, prioritising meaning preservation, clarity and minimal reformulation. Generator B uses RigoChat-7B-v2 with a more active Plain Language-oriented strategy, aimed at simplifying sentence structure, reducing lexical difficulty and improving directness while preserving the source meaning.

Generator C uses Gemini 2.5 Flash with an Easy-to-Read-oriented strategy, guided by the Spanish UNE 153101 recommendations and by a target linguistic profile close to CEFR A2. The CEFR A2 criterion was used as an operational linguistic-complexity target, not as evidence of actual reading comprehension. This third strategy therefore combines E2R-oriented constraints, short sentences, common vocabulary, explicit structure, visual segmentation when appropriate, and internal checks for meaning preservation and factual consistency.

Candidate diversification was intended to compare conservative, Plain Language-oriented and Easy-to-Read-oriented rewriting behaviours before selecting or refining the final output.

\paragraph{Candidate evaluation and routing.}

Each candidate is assessed by the candidate evaluator. The evaluator does not rely on a single quality score, but on groups of internal signals related to semantic and factual preservation, syntactic clarity, lexical simplicity, readability and robustness. These signals include checks for the preservation of numbers, dates, negations and conditions; sentence-length and splitting indicators; difficult-word and glossary-related indicators; readability-support signals; unsupported-content heuristics; and robustness checks for repetition or unexpected script changes.

The routing component applies ECA-style rules over these signals. Three main routing strategies are available. In V1, the best candidate is selected without modification. In V2, one candidate is selected as the base and minimally edited using another candidate as support. In V3, two candidates are merged only when they provide complementary improvements and no semantic-risk signal is detected. The router may also trigger targeted retries when semantic preservation, syntactic clarity, lexical simplification or E2R formatting is insufficient.

\paragraph{Controlled editing and final validation.}

The merger-editor is used only under controlled conditions. It receives the original text, the selected base candidate, an alternative candidate and a restricted set of permitted improvements. Its role is not to freely rewrite the text, but to make minimal edits or perform controlled fusion while preserving relevant information, numbers, dates, negations, conditions and domain-specific terms.

The final evaluator verifies the selected or edited output before submission. Critical signals related to factual consistency, semantic preservation and generation robustness are recalculated at this stage. The final decision is recorded together with the selected strategy, action and trace information. These traces were used during development for qualitative inspection and rule calibration, but no human intervention was used during official inference.

This workflow was used in both RUN1 and RUN2. The only architectural difference between the two submitted configurations was the activation of the lexical agent in RUN2, which introduced glossary-based and lexical-resource support before generation.

The workflow described in this section corresponds to the strategy followed for the MER-TRANS 2026 Spanish submissions. It should not be interpreted as the only possible strategy of the broader HULAT-UC3M accessible text generation infrastructure. Other configurations, prompts, routing rules, validation criteria and human-supervision mechanisms may be used in non-competition scenarios, depending on the domain, target users, available references and deployment constraints.

The signal taxonomy, ECA rule specification and threshold configuration are documented in the accompanying supplementary material.

\subsection{Baseline System}\label{subsec:baseline}

RUN3 was implemented as a linear generate--evaluate--regenerate baseline. It used RigoChat-7B-v2 with prompt-engineering strategies, post-processing and LoRA-based adaptation, a parameter-efficient fine-tuning technique that freezes the base model and injects trainable low-rank matrices into the model layers \cite{hu2021lora}.

The LoRA adapter was trained with public Spanish E2R material, mainly the Spanish Constitution in original and Easy-to-Read versions. The internal split contained 169 aligned articles: 126 for training and 43 for internal testing. During development, RigoChat-7B-v2 and Latxa-Qwen3-VL-8B-Instruct were compared under the same LoRA, prompting and quality-control conditions. Although Latxa obtained higher internal SARI and MeaningBERT scores in some configurations, RigoChat-7B-v2 was selected for RUN3 because it produced more stable outputs, with fewer repetitions and truncations.

The final RUN3 configuration used RigoChat-7B-v2 with the LoRA adapter, explicit A2/E2R-oriented prompts, controlled generation parameters and a regeneration prompt activated when internal quality criteria were not met. The generation parameters were set to favour controlled outputs: \texttt{max\_new\_tokens}=512, \texttt{temperature}=0.3, \texttt{top\_p}=0.85, \texttt{top\_k}=10, \texttt{repetition\_penalty}=1.25, \texttt{no\_repeat\_ngram\_size}=3 and \texttt{length\_penalty}=0.8. The quality-control module checked readability and semantic preservation using Fern\'andez-Huerta and MeaningBERT, and also detected critical errors such as repetition, truncation, language mixing, loss of numbers or dates, loss of negations and excessive information loss \cite{beauchemin2023meaningbert,fernandezhuerta1959lecturabilidad}. Regeneration was attempted up to two times.

\section{Experimental Setup}\label{sec:experimental_setup}

During development, the systems were not configured exclusively to maximise the official MER-TRANS scores. The official-style metrics were used as external benchmarking indicators, but no single metric was treated as a complete proxy for E2R quality. Accessibility-oriented simplification also requires checking whether the output preserves meaning, avoids unsupported additions, simplifies syntax and vocabulary, and remains readable for the intended audience.

To support this broader view, the HULAT-UC3M \textit{bilingual\_simplification\_metrics} framework was used as a development-time support tool \cite{hulat2026bilingualmetrics}. During calibration, when paired source--reference data were available, reference-based metrics such as SARI, BLEU and BERTScore were used to compare candidate behaviours and adjust prompts, routing rules and retry conditions. These metrics were not available during official test inference because the gold references were not provided to participants.

During official inference, routing decisions were mainly guided by reference-free internal signals computed from the source text and the candidate output. These signals were also informed by the empirical evidence base described in Section~\ref{subsec:resources}, which supported the definition of quality priorities, checklists and ECA-style control rules. During development and calibration, additional semantic and factual-consistency metrics available in the \textit{bilingual\_simplification\_metrics} framework, including MeaningBERT, AlignScore, QuestEval and SummaC, were inspected when the required inputs and model configurations were available \cite{hulat2026bilingualmetrics}.

For the MER-TRANS official inference setting, the internal signals used for routing were grouped into five operational categories:

\begin{itemize}
    \item Semantic and factual preservation: semantic-preservation support signals and checks for numbers, dates, negations, conditions and source-supported content.
    \item Readability: Fern\'andez-Huerta readability and related sentence-level indicators.
    \item Syntactic clarity: sentence length, punctuation complexity and splitting behaviour.
    \item Lexical simplicity: difficult-word ratio, glossary hits, unresolved difficult terms and lexical substitution behaviour.
    \item Robustness: repeated fragments, unexpected scripts and other generation errors.
\end{itemize}

These development-time metrics and inference-time signals were used to calibrate prompts, define retry conditions and route candidate outputs through ECA-style rules. For example, a candidate with adequate meaning preservation but overly long sentences could trigger a structural simplification action; a candidate with unresolved glossary terms could activate lexical support; and a candidate with factual-risk signals could be rejected or sent to a conservative retry.

The internal signals were therefore used to configure and control the system, not as alternative official scores. The official results reported in Section~\ref{sec:results} correspond only to the leaderboard metrics computed by the organizers.

\section{Results}\label{sec:results}

Table~\ref{tab:hulat2_results} isolates the official results of the three HULAT2-UC3M runs in order to compare the submitted configurations directly. Table~\ref{tab:leaderboard_context} then places these runs within the complete Spanish leaderboard.

\begin{table}[t]
\centering
\caption{Official MER-TRANS Spanish results for the three HULAT2-UC3M configurations.}
\label{tab:hulat2_results}
\begin{tabular}{lcccc}
\toprule
\textbf{Run} & \textbf{BLEU-Orig} & \textbf{BLEU-Gold} & \textbf{SARI} & \textbf{BERTScore} \\
\midrule
RUN1 & 35.5730 & 17.5715 & 44.0543 & 0.9263 \\
RUN2 & 34.1564 & 16.1484 & 43.1049 & 0.9248 \\
RUN3 & 2.4037 & 5.3373 & 38.5136 & 0.9114 \\
\bottomrule
\end{tabular}
\end{table}

RUN1 obtained the best official SARI score among the three submitted HULAT2 runs. It improved over RUN3 by 5.5407 SARI points, which corresponds to a relative improvement of approximately 14.39\% over the baseline SARI score. RUN2 also improved over RUN3 by 4.5913 SARI points. These results suggest that, in this setting, the signal-guided multi-agent workflow was more effective than the linear generate--evaluate--regenerate baseline.

RUN2, which activated the lexical agent, obtained a slightly lower official SARI score than RUN1. This suggests that the lexical support introduced in this configuration did not improve the official reference-based ranking. However, this result should not be interpreted as a general conclusion against lexical resources. Rather, it indicates that glossary-based or domain-informed lexical substitutions must be carefully balanced with semantic preservation, contextual adequacy and alignment with the reference simplifications.

RUN3 obtained lower official scores and served mainly as an exploratory baseline. Its very low BLEU-Orig score indicates that the baseline outputs diverged strongly from the source wording, but this divergence was not compensated by higher similarity to the gold E2R reference. This suggests that the baseline sometimes rewrote too aggressively or produced outputs that were not sufficiently aligned with the expected simplification style.

\begin{table*}[t]
\centering
\caption{Official Spanish leaderboard for MER-TRANS 2026, sorted by SARI. HULAT2-UC3M runs are highlighted in bold.}
\label{tab:leaderboard_context}
\small
\begin{tabular}{rllrrrr}
\toprule
\textbf{Rank} & \textbf{Team} & \textbf{Run} & \textbf{BLEU-Orig} & \textbf{BLEU-Gold} & \textbf{SARI} & \textbf{BERTScore} \\
\midrule
1  & ClearText   & RUN1 & 18.0013  & 15.4483 & 47.0168 & 0.9220 \\
2  & VICOMTECH   & RUN3 & 20.7912  & 14.1086 & 45.5843 & 0.9241 \\
3  & HULAT1      & RUN2 & 33.1815  & 17.7135 & 45.0182 & 0.9272 \\
4  & HULAT1      & RUN3 & 31.5608  & 17.1627 & 44.7465 & 0.9272 \\
5  & VICOMTECH   & RUN1 & 12.4857  & 9.3316  & 44.1880 & 0.9114 \\
\textbf{6}  & \textbf{HULAT2} & \textbf{RUN1} & \textbf{35.5730} & \textbf{17.5715} & \textbf{44.0543} & \textbf{0.9263} \\
7  & HULAT1      & RUN1 & 37.1581  & 18.2081 & 43.3874 & 0.9276 \\
8  & BASELINE1   & RUN1 & 18.9491  & 10.8850 & 43.2873 & 0.9130 \\
\textbf{9}  & \textbf{HULAT2} & \textbf{RUN2} & \textbf{34.1564} & \textbf{16.1484} & \textbf{43.1049} & \textbf{0.9248} \\
10 & VICOMTECH   & RUN2 & 32.0715  & 14.7485 & 42.1724 & 0.9246 \\
11 & HumanAI-UCM & RUN2 & 26.9792  & 11.8405 & 41.6196 & 0.9111 \\
12 & FACILE      & RUN1 & 3.0009   & 2.8644  & 41.6129 & 0.8783 \\
13 & NIL\_UCM    & RUN2 & 11.3818  & 4.7980  & 40.5175 & 0.9026 \\
14 & HumanAI-UCM & RUN1 & 38.3402  & 15.0830 & 40.4163 & 0.9209 \\
15 & NIL\_UCM    & RUN1 & 6.0574   & 2.5775  & 39.6873 & 0.8780 \\
\textbf{16} & \textbf{HULAT2} & \textbf{RUN3} & \textbf{2.4037} & \textbf{5.3373} & \textbf{38.5136} & \textbf{0.9114} \\
17 & BASELINE2   & RUN1 & 29.0913  & 9.3904  & 37.1761 & 0.9019 \\
18 & HumanAI-UCM & RUN3 & 45.0236  & 14.5298 & 36.2450 & 0.9125 \\
19 & DoNothing   & RUN1 & 100.0000 & 23.2176 & 13.8687 & 0.9297 \\
\bottomrule
\end{tabular}
\end{table*}

Table~\ref{tab:leaderboard_context} reports the complete official Spanish leaderboard, sorted by SARI. HULAT2-RUN1 ranked 6th out of 19 Spanish submissions, HULAT2-RUN2 ranked 9th, and HULAT2-RUN3 ranked 16th. Therefore, HULAT2-RUN1 was competitive within the Spanish track, although it was not the top-ranked system. The main claim of this paper is not state-of-the-art performance, but the comparative finding that the signal-guided multi-agent architecture outperformed the RigoChat-based linear baseline submitted by the same team.

The complete leaderboard also illustrates why BLEU-Orig and BERTScore must be interpreted with caution in this task. The DoNothing baseline preserves the original text and therefore obtains the highest BLEU-Orig and a high BERTScore, but it obtains the lowest SARI score. This confirms that source overlap and semantic similarity alone do not measure whether a text has actually been simplified.

\section{Error Analysis}\label{sec:error_analysis}

A preliminary error-oriented analysis was conducted based on the official scores, the comparison between the submitted configurations, and the behaviour observed during development. Since the official gold references were not available for detailed segment-level inspection at the time of writing, this section should be interpreted as a preliminary analysis rather than as a complete error annotation.

The results point to three main areas for further inspection. First, the lower performance of RUN3 suggests that the baseline may have produced overly aggressive rewritings, reduced reference alignment or insufficient control over information preservation. This is consistent with the large drop in BLEU-Orig and the lower BLEU-Gold and SARI scores observed for this run.

Second, the difference between RUN1 and RUN2 suggests that lexical support requires careful calibration. Glossary-based explanations or lexical substitutions may be useful from an accessibility perspective, but they can also increase divergence from the expected reference if they introduce unnecessary paraphrases, alter terminology in context or expand information beyond what the reference expresses.

Third, the leaderboard confirms that high source overlap and high semantic similarity are not sufficient indicators of successful simplification. The DoNothing baseline shows that preserving the original text can lead to high BLEU-Orig and BERTScore values while failing to simplify the input. This supports the need to complement official metrics with qualitative and accessibility-oriented error analysis.

Future error analysis should combine local segment-level inspection with document-aware assessment. Segment-level inspection is needed to detect local errors such as loss of numbers, dates, negations or conditions, unsupported additions, excessive rewriting and insufficient simplification. Document-aware assessment is needed to analyse consistency across segments, terminology stability, the presence of repeated or missing explanations, the coherence of the reconstructed document, and the homogeneity of the E2R style.

\section{Conclusions}\label{sec:conclusions}

This paper has presented the HULAT2-UC3M participation in the Spanish track of MER-TRANS 2026. Three fully automatic runs were submitted: two configurations of a LangGraph-based multi-agent workflow and one linear generate--evaluate--regenerate baseline.

RUN1 was the highest-scoring HULAT2 run according to the official SARI score, with 44.0543 points, followed by RUN2 with 43.1049 and RUN3 with 38.5136. These results suggest that, for this task, candidate diversification, automatic evaluation signals, ECA routing and controlled editing were more effective than the linear baseline. RUN1 and RUN2 differed only in the activation of the lexical agent, so the lower score of RUN2 should be interpreted within the official evaluation setting, not as a general conclusion against lexical resources. Rather, it shows that lexical support requires careful calibration: it may support user-centred explanations, terminology control, and domain-specific accessibility, but it does not automatically improve performance on reference-based metrics.

Beyond leaderboard performance, the HULAT2-UC3M approach is oriented towards increasing the robustness and controllability of accessible text generation systems. The submitted workflow was designed not only to generate simplified text, but also to monitor semantic preservation, factual consistency, readability, lexical difficulty and generation errors through internal signals and traceable routing decisions.

Future work will focus on document-aware error analysis, combining local segment-level inspection with assessment of consistency, terminology stability, coherence and E2R style across reconstructed documents. Expert review, user-centred validation and the integration of Human-in-the-Loop and Human-on-the-Loop mechanisms will also be explored in non-competitive deployments. This will make it possible to complement automatic metrics with accessibility-oriented evidence about comprehension, adequacy and usefulness for target users.

\section*{Declaration on Generative AI}

Generative AI tools were used to improve the wording in English and to edit some LaTeX text. The authors checked and edited the final version. The authors take full responsibility for the content of the paper.

\section*{Availability of Code and Supplementary Material}

The implementation code associated with the HULAT2-UC3M MER-TRANS 2026 submissions is available in the task-associated GitHub repository: \url{https://github.com/hulat-group/mertrans_2026}. The accompanying reproducibility material, including the signal taxonomy, ECA rule specification, routing logic, validation signals and configuration documentation, is archived in Zenodo \cite{moreno2026aigovaccess}.

\section*{Acknowledgments}
This work has been supported by grant PID2023-148577OB-C21 (Human-Centered AI: User-Driven Adapted Language Models-HUMAN\_AI) by MICIU/AEI/10.13039/501100011033 and by FEDER/UE.

\bibliography{references}

\appendix
\section{Public Resources Used for Calibration and Lexical Support}
\label{app:public_resources}

The following public resources were used during system calibration, prompt refinement or glossary construction:

\begin{itemize}
    \item Plena Inclusión España, The Spanish Constitution in Easy-to-Read Format, Easy-to-Read publication, \url{https://www.plenainclusion.org/publicaciones/buscador/
la-constitucion-espanola-en-lectura-facil/}.
    \item Easy-to-Read citizen participation guide from the Spanish transparency portal: \url{https://transparencia.gob.es/content/dam/transparencia_home/multimedia/pdf/guias-de-lectura-facil/Guia_participacion_ciudadana_v3_sinplanes_n.pdf}
    \item Citizen participation material available through SlideShare: \url{https://es.slideshare.net/slideshow/lectura-facil-participacin-ciudadanapdf/255259357}
    \item Accessible citizen participation manual from CARM: \url{https://participa.carm.es/documents/5690123/6505112/C\%C3\%B3mo+participar.+Manual+Accesible/b6624045-65ce-43c3-8ab8-ece6f8852d23}
    \item Plena Inclusi\'on opposition-exam materials: \url{https://www.plenainclusion.org/coleccion/temario-de-oposicion/}
    \item Easy-to-Read guide for electoral participation: \url{https://mivotocuenta.es/wp-content/uploads/2019/02/Guia-para-participar-en-las-elecciones-en-Lectura-F\%C3\%A1cil.pdf}
    \item SimpMedLexSp \cite{campillos2022clara} and related Spanish medical simplification resources.
\end{itemize}

\end{document}